\pdfoutput=1

\documentclass[11pt]{article}

\usepackage[final]{EMNLP2023}

\usepackage{times}
\usepackage{latexsym}
\usepackage{tcolorbox}
\usepackage{fancyvrb}
\usepackage{amsmath}
\tcbuselibrary{skins}
\usepackage[T1]{fontenc}

\usepackage[utf8]{inputenc}
\usepackage{microtype}
\usepackage{booktabs}
\usepackage{subcaption}
\usepackage{inconsolata}
\usepackage{pgfplots}
\pgfplotsset{compat=1.17}
\usepackage{algorithm}
\usepackage{algpseudocode}
\usepackage{microtype}

\usepackage{inconsolata}

\usepackage{graphicx}
\title{TreePrompt: Leveraging Hierarchical Few-Shot Example Selection for Improved English-Persian and English-German Translation}

\author{
  Ramtin Kakavand, Ebrahim Ansari  \\
  Department of Computer Science \\
  Institute for Advanced Studies in Basic Sciences \\
  Zanjan, Iran \\
  \texttt{\{r.kakavand, ansari\}@iasbs.ac.ir}
}

\begin{document}
\maketitle

\begin{abstract}
Large Language Models (LLMs) have consistently demonstrated strong performance in machine translation, especially when guided by high-quality prompts. Few-shot prompting is an effective technique to improve translation quality; however, most existing example selection methods focus solely on query-to-example similarity and do not account for the quality of the examples. In this work, we propose TreePrompt, a novel example selection approach that learns LLM preferences to identify high-quality, contextually relevant examples within a tree-structured framework. To further explore the balance between similarity and quality, we combine TreePrompt with K-Nearest Neighbors (K-NN) and Adaptive Few-Shot Prompting (AFSP). Evaluations on two language pairs—English–Persian (MIZAN) and English–German (WMT19)—show that integrating TreePrompt with AFSP or Random selection leads to improved translation performance.
\end{abstract}

\section{Introduction}

Large language models (LLMs) demonstrate remarkable capabilities in machine translation (MT) tasks\citep{bawden2023investigating}. Significant advancements have been made in these models, particularly with architectures such as ChatGPT and GPT-4 \citep{ouyang2022training,hendy2023good, zhang2023machine}.

Few-shot prompting plays a crucial role in improving translation quality and enhancing a model’s ability to generalize with limited data \citep{vilar2022prompting,chen2023unleashing}. By selecting appropriate examples, we can effectively unlock the translation potential of LLMs \citep{tang2025adaptive}. Although numerous example selection strategies have been proposed, most primarily focus on the similarity between the examples and the input query.
\begin{figure}[htbp]
  \centering
  \includegraphics[width=0.5\textwidth]{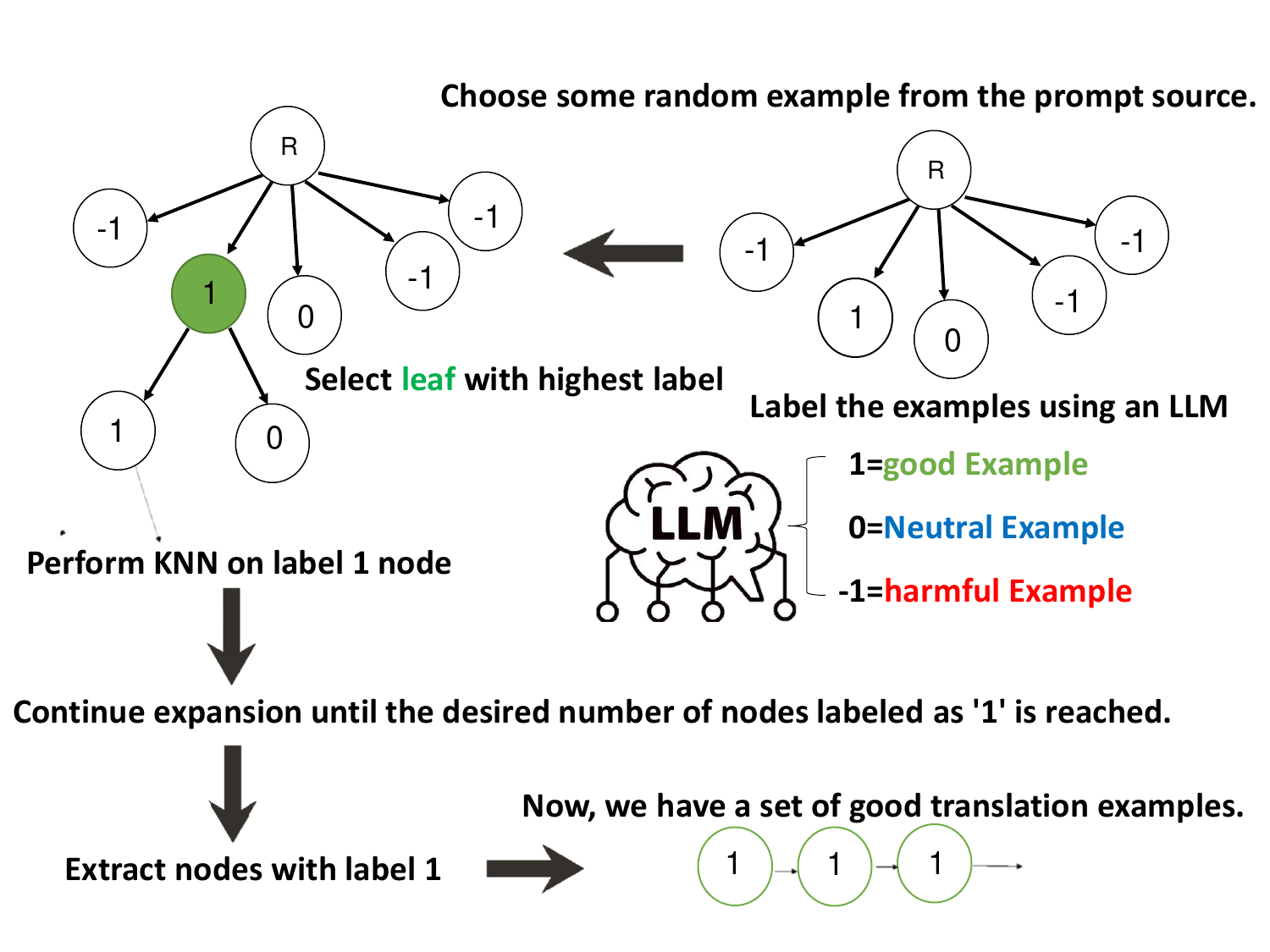}
  \caption{An overview of the proposed Tree-based example selection approach for few-shot translation prompting }
  \label{fig:vector-diagram}
\end{figure}

To address the limitations of existing few-shot prompting techniques—such as the irrelevance of randomly sampled examples and the similarity-based selection methods used in models like KNN and AFSP \citep{tang2025adaptive}, which often overlook example quality—we propose a novel and adaptive approach called TreePrompt. Unlike traditional methods, TreePrompt dynamically incorporates the translation preferences of the LLM by starting with a set of randomly selected examples that are labeled by the LLM as positive, neutral, or negative.

Following this labeling, TreePrompt iteratively refines the prompt by expanding only the most promising (i.e., positively labeled) examples using a KNN algorithm applied over RoBERTa embeddings, effectively constructing a tree-like structure of semantically related examples. Rather than relying solely on similarity, TreePrompt leverages the LLM’s preferences to select high-quality, impactful few-shot examples, resulting in more effective and model-aligned prompts. Figure~\ref{fig:vector-diagram} illustrates this tree-based selection strategy for enhancing few-shot prompting in machine translation.

We evaluate TreePrompt on two datasets: the MIZAN Persian–English parallel corpus \citep{kashefi2018mizan} and the WMT19 English–German dataset \citep{wmt19translate}. On the WMT19 dataset, we test one LLM (DeepSeek) with eight different selection techniques, while on the MIZAN dataset, we compare nine selection techniques across various LLMs. To assess both semantic similarity and translation quality, we combine TreePrompt with Adaptive Few-Shot Prompting (AFSP) and K-Nearest Neighbors (KNN). Experimental results show that TreePrompt integrated with AFSP or with Random sampling plus a reranker consistently outperforms baseline methods across all evaluation metrics, with the largest improvements observed in COMET scores. These findings highlight TreePrompt’s potential as a flexible and effective framework for selecting high-quality examples, particularly in low-resource machine translation scenarios.

Our primary contribution is the integration of LLM preferences into few-shot example selection. We achieve this by constructing a tree-structured selection process that begins with a set of randomly chosen examples and incrementally incorporates the most promising ones based on LLM feedback. Semantic relationships among these high-quality examples are modeled using KNN. Furthermore, to balance example similarity and quality, we integrate TreePrompt with two similarity-driven methods—KNN and AFSP—resulting in more efficient and adaptive prompting strategies.
\section{Related Work}
Enhancing the translation performance of large language models (LLMs) has been extensively studied across various domains. Tang et al. \citep{tang2025adaptive} provide a comprehensive summary of this work. Key research directions include document-level translation \citep{wang2022understanding,karpinska2023large}, multilingual translation \citep{hendy2023good,jiao2023chatgpt}, self-refining methods \citep{feng2024tear}, and agentic pipelines \citep{wu2024perhaps,guo2024sillm}. Low-resource translation has also attracted significant attention \citep{jiao2023chatgpt,bawden2023investigating}, along with the design of effective prompt templates \citep{zhang2023machine,zhang2023prompting,jiao2023chatgpt} and demonstration selection strategies for in-context learning \citep{garcia2023unreasonable,merx2024low,jiang2024convergences}.
The emergence of large language models (LLMs) in recent years has significantly enhanced the overall performance of machine translation (MT), especially through prompting methods such as few-shot learning. In this approach, LLMs are given a few source-target translation pairs as context to generate translations for new inputs.\citet{vilar2022prompting}, \citet{briakou2023searching}, and \citet{chowdhery2023palm} demonstrated that PaLM can achieve promising machine translation results when provided with in-context examples. \citet{pourkamali2024machine} compared LLMs in few-shot settings for English–Persian and English–Russian translation, concluding that the number of examples is not a critical factor for improving performance, especially for resource-poor target languages. Instead, \citet{vilar2022prompting} emphasized that example quality is more important than query–example similarity. \citet{zhang2023prompting} found that using suboptimal examples can negatively impact translation quality. Moreover, \citet{zhu2023multilingual} and \citet{guo2024teaching} showed that including cross-lingual exemplars can enhance performance in low-resource translation tasks.
Various prompt selection methods have been developed to enhance few-shot machine translation (MT). One widely used approach is K-Nearest Neighbors (KNN) \citep{khandelwal2020nearest}, which retrieves examples most similar to the test input based on sentence embeddings such as SBERT or RoBERTa 
\citep{roberta, sbert, Chehreh2024}. While KNN-based selection is efficient and adaptive to input, it remains purely embedding-driven and can sometimes misalign with the internal preferences of large language models (LLMs).
Another notable method is Adaptive Few-Shot Prompting (AFSP). \citep{tang2025adaptive}, which selects prompt samples based on their contextual fluency and accuracy using a hybrid retrieval module. Although more adaptive than KNN, AFSP lacks a structural expansion mechanism and depends on scoring a fixed set of candidate examples.
Previous work has highlighted the importance of example selection and model-guided prompting in machine translation (MT). Our proposed TreePrompt method offers a novel solution by bridging the gap between semantic similarity and model-driven preferences through a label-guided expansion scheme with built-in constraints. This approach generates higher-quality and more interpretable prompts, which is especially beneficial in low-resource settings such as Persian–English machine translation.

\begin{figure}[ht]
\centering
\begin{minipage}{0.98\linewidth}
\tiny
\begin{tcolorbox}[
    enhanced,
    colback=gray!5,
    colframe=black,
    boxrule=0.3pt,
    sharp corners,
    left=4pt, right=4pt, top=4pt, bottom=4pt,
    width=\linewidth,
    fonttitle=\bfseries,
    title=Prompt for Evaluating Translation Example
]
Evaluate the following translation example for use in a 5-shot few-shot translation task.

\textbf{Source Language}: [source language] \\
\textbf{Target Language}: [target language]

\textbf{Source Text Set (SRT)}: "[SRT]"

\textbf{Translation Example:} \\
\textbf{Source}: "[ST]" \\
\textbf{Target}: "[TT]"

Assign a score based on these criteria: \\
1 — Improves translation quality \\
0 — Neutral effect \\
-1 — Degrades translation quality

\textbf{Return only the score: 1, 0, or -1.}
\end{tcolorbox}
\caption{Compact prompt used for scoring translation examples}
\label{fig:compact-prompt}
\end{minipage}
\end{figure}

\section{Proposed Method }
Our method is based on the premise that the quality of selected examples is more important than their similarity to the query \citep{vilar2022prompting}. We implement our approach using a tree structure that incorporates the preferences of the LLM during example selection. In the initial step, we randomly select several examples from the prompt source and pick a test sentence from the query set. We then ask the LLM to label these examples according to the following criteria:
\begin{figure}[h]
\centering
\begin{algorithm}[H]
\caption{TreePrompt: Tree-Based Example Selection Algorithm}
\begin{algorithmic}[1]
\Require Query Set $Q$, Prompt Source Corpus $P$
\Ensure A refined subset of examples $E'$ for few-shot prompting
\State Initialize candidate set $E \subset P$ by randomly sampling examples
\For{each example $e \in E$}
    \State Query the LLM to assign a label $l_e \in \{-1, 0, 1\}$ indicating translation quality
\EndFor
\While{the number of positively labeled examples ($l_e = 1$) is below a desired threshold}
    \State Select the most promising leaf node (examples with label 1 preferred, then 0, then -1)
    \State Use $k$-Nearest Neighbors (based on RoBERTa embeddings) to retrieve semantically similar examples from $P$
    \State Query the LLM to label the new examples using the updated source set
    \State Add newly labeled examples to the candidate set $E$
\EndWhile
\State \Return Final set $E' \subseteq E$ containing the best-labeled examples for prompting
\end{algorithmic}
\end{algorithm}
\caption{TreePrompt procedure: Iteratively selects and expands high-quality examples based on LLM preference and KNN similarity.}
\label{fig:TreePrompt-algorithm}
\end{figure}

\begin{itemize}
    \item \textbf{1}: Assign label 1 if this example provides valuable insight for translating the test sentence and is suitable for few-shot prompting.
    \item \textbf{0}: Assign label 0 if this example neither improves nor degrades the translation quality of the test sentence when used in few-shot prompting.
    \item \textbf{-1}: Assign label -1 if this example, when used in few-shot prompting, degrades the translation quality of the test sentence.
\end{itemize}

Inspired by sentiment analysis\citep{panneerselvamthumbs}, our labeling method uses the prompt shown in Figure ~\ref{fig:compact-prompt}. After labeling, we expand the set of high-quality examples by applying K-Nearest Neighbors (K-NN) to the best-labeled example. Specifically, the source sentence of the top example serves as the new query for K-NN retrieval, which returns semantically similar candidates. This process is repeated until the desired number of high-quality examples is collected.

As previously mentioned, this method operates within a tree structure, where each node represents an example labeled as 1, 0, or -1. The procedure starts by randomly extracting several examples from the prompt source along with one test sentence from the test set. Next, these examples are labeled according to the criteria (1, 0, -1). Then, the leaf node with the best label is selected and expanded using K-NN. Figure~\ref{fig:TreePrompt-algorithm} illustrates the TreePrompt procedure.

Performing this process for every test sentence would be computationally expensive and inefficient. Therefore, we find high-quality examples once for a set of test sentences rather than individually for each one. To ensure diversity, the test sentence used for labeling is changed in each iteration—meaning the LLM labels examples based on a different randomly selected test sentence from the test set each time.
\subsection{Complexity Analysis}
\label{sec:complexity}

We analyze the computational complexity of the \textbf{TreePrompt} algorithm in terms of both time and space.

Let:
\begin{itemize}
    \item $n$ be the size of the prompt source corpus $P$,
    \item $m$ be the number of examples initially sampled from $P$,
    \item $k$ be the number of neighbors retrieved per iteration,
    \item $d$ be the dimensionality of the embedding space (e.g., RoBERTa),
    \item $t$ be the number of while-loop iterations until the desired number of positively labeled examples is collected,
    \item $c$ be the cost of querying the LLM to label a single example.
\end{itemize}

\subsection{Time Complexity}
The algorithm consists of the following main steps:

\begin{enumerate}
    \item \textbf{Initialization}: Randomly sampling $m$ examples from $P$ takes $\mathcal{O}(m)$ time.
    \item \textbf{Initial Labeling}: Querying the LLM to label the initial $m$ examples incurs a cost of $\mathcal{O}(m \cdot c)$.
    \item \textbf{While Loop} (repeated for $t$ iterations):
    \begin{itemize}
        \item \textbf{Leaf Selection}: Selecting the most promising node from the current candidate set $E$ takes $\mathcal{O}(|E|) = \mathcal{O}(m + t \cdot k)$ time.
        \item \textbf{KNN Retrieval}: Using brute-force KNN based on $d$-dimensional embeddings incurs a cost of $\mathcal{O}(n \cdot d)$. This can be improved using approximate methods (e.g., FAISS).
        \item \textbf{LLM Labeling of New Examples}: Querying $k$ new examples takes $\mathcal{O}(k \cdot c)$ time.
        \item \textbf{Update Candidate Set}: Adding new examples to $E$ takes $\mathcal{O}(k)$ time.
    \end{itemize}
\end{enumerate}

Therefore, the overall time complexity of the algorithm is:
\[
\mathcal{O}(m \cdot c + t \cdot (n \cdot d + k \cdot c))
\]

If an approximate nearest neighbour (ANN) search algorithm is used instead of brute-force KNN, the complexity can improve to:
\[
\mathcal{O}(m \cdot c + t \cdot (\log n + k \cdot c))
\]

\subsection{Space Complexity}
The space requirements are as follows:
\begin{itemize}
    \item $\mathcal{O}(n \cdot d)$ to store embeddings for the entire corpus $P$,
    \item $\mathcal{O}(m + t \cdot k)$ for maintaining the candidate set $E$.
\end{itemize}

Hence, the total space complexity is:
\[
\mathcal{O}(n \cdot d + m + t \cdot k)
\]

\subsection{Summary}
The main costs in the TreePrompt algorithm are LLM calls and similarity search over the prompt corpus. Utilizing KNN in a computationally efficient way, as well as batching or caching LLM requests, can significantly reduce real-world runtime.

\section{Experimental Setup }
This section describes the model configurations, sampling methods, evaluation criteria, and datasets used in our experiments.

\subsection{Language Models}

We evaluated our approach using four large language models:

    GPT-4o: The latest multimodal model developed by OpenAI, known for its superior performance in multilingual tasks, including machine translation \citep{openai2024gpt4o}.

    GPT-3.5-turbo: An efficient and faster variant of GPT-3.5 that demonstrates strong translation capabilities \citep{openai2023gpt35}.

    DeepSeek-V2: An open-source competitive large language model that produces high-quality outputs for machine translation tasks \citep{deepseek2024v2}.

\subsection{Tree Expansion via KNN-RoBERTa}

To augment the tree structure in our proposed TreePrompt method, we utilize the K-Nearest Neighbors (KNN) algorithm applied to sentence embeddings generated by the RoBERTa model. Starting from a labeled example, KNN-RoBERTa allows us to retrieve semantically similar samples from the prompt source corpus, which are subsequently considered for further labeling and integration into the tree.

We selected RoBERTa due to its superior ability to capture semantic similarity compared to other embedding models such as BERT or Sentence-BERT. Empirical evaluations demonstrate that retrieval based on RoBERTa embeddings consistently yields higher-quality example expansions, thereby improving the effectiveness of few-shot prompts in downstream translation tasks \citep{reimers2019sentence}. The embedding strategy employed here plays a critical role in guiding the LLM toward selecting more contextually relevant translation examples.

\begin{figure}[ht]
\centering
\begin{tcolorbox}[colback=gray!5, colframe=black!60, width=0.95\linewidth, arc=2mm, boxrule=0.5pt, fontupper=\small]
You are a professional translator. I will give you one or more examples of text fragments, where the first one is in \texttt{\{src lang\}} and the second one is the translation of the first fragment into \texttt{\{tgt lang\}}. These sentences will be displayed below.\\

1. \texttt{\{src lang\}} text: \texttt{\{src demo 1\}}\\
\texttt{\{tgt lang\}} translation: \texttt{\{tgt demo 1\}}\\
2. \texttt{\{src lang\}} text: \texttt{\{src demo 2\}}\\
\texttt{\{tgt lang\}} translation: \texttt{\{tgt demo 2\}}\\
3. \texttt{\{src lang\}} text: \texttt{\{src demo 3\}}\\
\texttt{\{tgt lang\}} translation: \texttt{\{tgt demo 3\}}\\
...\\
k. \texttt{\{src lang\}} text: \texttt{\{src demo k\}}\\
\texttt{\{tgt lang\}} translation: \texttt{\{tgt demo k\}}\\

After the example pairs, I will provide a sentence in \texttt{\{src lang\}} and would like you to translate it into \texttt{\{tgt lang\}}. Please provide only the translation result without any additional comments, formatting, or chat content. Translate the text from \texttt{\{src lang\}} to \texttt{\{tgt lang\}}.
\end{tcolorbox}
\caption{Prompt used for few-shot translation}
\label{fig:prompt}
\end{figure}

\subsection{Few-shot Configuration}

We adopted a 5-shot setup, as increasing beyond five examples can degrade translation quality \citep{peng2023towards, garcia2023unreasonable}, and this number balances translation accuracy with computational efficiency. Few-shot samples were selected using four methods: Random, KNN, AFSP, and our proposed Tree-based approach.

\subsection{Implementation of Adaptive Few-Shot Prompting and K-NN}

For Adaptive Few-Shot Prompting (AFSP), a hybrid retrieval step is employed to select examples \citep{tang2025adaptive}. When working with sparse embeddings, we utilize the paraphrase-MiniLM-L6-v2 model, whereas for dense embeddings, the all-MiniLM-L6-v2 model is used. To handle multi-vector embeddings, the multi-qa-mpnet-base-dot-v1 model is applied. The same large language model (LLM) serves both for translation and reranking. Retrieval weights are initialized as 0.4, 0.4, and 0.2 for the three embedding types respectively, a configuration that has demonstrated superior performance \citep{tang2025adaptive}.

For the K-Nearest Neighbors (K-NN) method, we utilize the K-NN RoBERTa model as described by \citet{vilar2022prompting}.
\subsection{Dataset}

We conducted all experiments using the MIZAN corpus \citep{kashefi2018mizan}, a high-quality Persian–English parallel dataset specifically designed for machine translation research. MIZAN is particularly valuable for evaluating translation systems involving Persian, a low-resource language that typically lacks large-scale, well-aligned parallel corpora. Due to its linguistic quality and alignment accuracy, MIZAN serves as a reliable benchmark for Persian–English MT tasks.

Since this study is a preliminary investigation, we selected a representative sample from the dataset: 9,000 sentence pairs were used as the prompt source corpus, while 520 sentence pairs were set aside as a test set for evaluating our few-shot prompting methods.

Additionally, we evaluated our approaches on the WMT19 English–German dataset \citep{wmt19translate}, using 9,000 sentence pairs as the prompt source and 500 sentences as the test set. These experiments were conducted exclusively with the DeepSeek model.

\subsection{Evaluation Metrics}

The machine translation outputs were evaluated using the following automatic metrics. Among these, COMET was chosen as the primary metric due to its strong correlation with human judgment:

    BLEU: Measures n-gram overlap between the machine-generated translation and a human reference. It is widely used in MT research despite its limitations in capturing semantic equivalence \citep{papineni2002bleu}.

    CHRF: A character-level F-score metric that captures morphological and orthographic differences, making it particularly suitable for morphologically rich languages like Persian \citep{popovic2015chrf}.

    COMET: A neural-based metric that predicts translation quality by comparing the candidate translation with both the source and reference sentences using pretrained language models. COMET has demonstrated strong correlation with human judgments and is used as the primary evaluation metric in this study \citep{rei2020comet}.

    BERTScore: Calculates semantic similarity using contextual embeddings from BERT, capturing nuanced meaning beyond surface-level overlap \citep{zhang2019bertscore}.

\subsection{Prompting Used for Few-Shot}

In our approach, the example selection phase is separated from the translation phase to enable precise evaluation of the selection methods. We adopt the same prompt structure for few-shot translation as introduced by \citet{tang2025adaptive}, applying it consistently across KNN, Random, and AFSP methods. The prompt used during the translation phase is illustrated in Figure~\ref{fig:prompt}.

\section{ Results}
\label{sec:bibtex}

We compare nine different methods: Zero-shot, Random, KNN, and AFSP individually, as well as their hybrids combined with our proposed Tree-based example selection—namely, Tree+Random, Tree+KNN, TreePrompt+AFSP, TreePrompt+Random+Reranker, and TreePrompt+KNN+Reranker. In all “Tree+” variants, we first apply our TreePrompt approach to the prompt source corpus to extract high-quality examples. We then apply the respective selection method (e.g., AFSP or KNN) on this pre-filtered set. For the reranking variants, the same LLM is used as a reranker to re-rank and score the final examples.

The outputs of these methods are presented in Figure~\ref{fig:llm-comet-compare}, comparing results across three different LLMs: GPT-4o \citep{openai2024gpt4o}, GPT-3.5-Turbo \citep{openai2023gpt35}, and DeepSeek \citep{deepseek2024v2}.
 
\subsection{Discussion}

The results show that COMET scores are negative across all methods, reflecting the challenges posed by Persian as a low-resource language for large language models. For the GPT-4o model, we extracted 144 and 324 high-quality translation examples using our tree-based selection criterion to examine how performance varies with the number of high-quality examples. The best COMET score (i.e., the least negative) was achieved by the \textit{TreePrompt-324 + AFSP} method. Conversely, the zero-shot configuration showed the worst performance, yielding the lowest COMET score among all methods tested.

In the case of GPT-3.5 Turbo, the Random with Reranker method outperformed other approaches based on COMET scores, though its overall performance remained inferior to that of GPT-4o. For the DeepSeek model, the highest COMET score was obtained by the \textit{TreePrompt + Random + Re-ranker} method.

Additionally, GPT-4o exhibited a more conservative labeling behavior, being more selective in assigning the top label (1) to translation samples.

This analysis suggests that GPT-4o employs more rigorous evaluation criteria when assessing translation quality. In contrast, DeepSeek and GPT-3.5 Turbo exhibit a more lenient scoring bias, showing greater willingness to assign a label of 1. As a result, the proportion of examples labeled as -1 or 0 is significantly higher for DeepSeek and GPT-3.5 Turbo compared to GPT-4o, indicating a less selective evaluation threshold in these models.

Table~\ref{tab:English-German results} presents the results on the WMT19 dataset, where K-NN achieves the highest COMET score among methods, although zero-shot performs best on other metrics. Across all language models, the Tree-based method proves effective for selecting high-quality translation examples.

\begin{figure}[ht]
\begin{minipage}[t]{0.48\textwidth}
\centering
\begin{tikzpicture}
\begin{axis}[
    width=6.5cm,
    height=5.5cm,
    xlabel={},
    ylabel={COMET},
    xtick=data,
    xticklabels={
        Zero-shot,
        Random,
        KNN,
        AFSP,
        TreePrompt+Random,
        TreePrompt+KNN,
        TreePrompt+AFSP,
        Random+Rerank,
        KNN+Rerank,
        TreePrompt+Random+Rerank,
        TreePrompt+KNN+Rerank
    },
    xticklabel style={rotate=45, font=\scriptsize, anchor=east},
    ylabel style={font=\scriptsize},
    xlabel style={font=\scriptsize},
    title={COMET Comparison},
    title style={font=\footnotesize},
    legend style={font=\tiny},
    tick label style={font=\tiny},
    ymajorgrids=true,
    grid style=dashed,
    legend pos=south east
]

\addplot+[mark=*, color=blue] coordinates {
    (0,-0.2334)
    (1,-0.1811)
    (2,-0.1859)
    (3,-0.1581)
    (4,-0.1963)
    (5,-0.1860)
    (6,-0.1839)
    (7,-0.1856)
    (8,-0.1760)
    (9,-0.1817)
    (10,-0.1758)
};
\addlegendentry{GPT-4o}

\addplot+[mark=square*, color=red] coordinates {
    (0,-0.3879)
    (1,-0.3600)
    (2,-0.3653)
    (3,-0.3361)
    (4,-0.3522)
    (5,-0.3455)
    (6,-0.3437)
    (7,-0.3279)
    (8,-0.3636)
    (9,-0.3459)
    (10,-0.3342)
};
\addlegendentry{GPT-3.5 Turbo}

\addplot+[mark=triangle*, color=green!70!black] coordinates {
    (0,-0.2037)
    (1,-0.1674)
    (2,-0.1943)
    (3,-0.1512)
    (4,-0.1492)
    (5,-0.1579)
    (6,-0.1701)
    (7,-0.1424)
    (8,-0.1617)
    (9,-0.1424)
    (10,-0.1617)
};
\addlegendentry{DeepSeek}

\end{axis}
\end{tikzpicture}
\caption{Compact COMET score plot across methods.}
\label{fig:comet-small}
\end{minipage}
\end{figure}
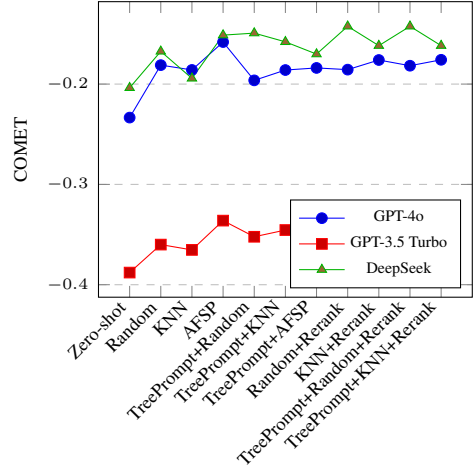

Figure~\ref{fig:comet-small} illustrates the COMET performance of each LLM. The results show that DeepSeek consistently outperforms GPT-4o across all prompting techniques, while GPT-4o outperforms GPT-3.5 Turbo.

\begin{figure}[ht]
\centering

\begin{subfigure}[b]{\linewidth}
\centering
\caption{GPT-4o}
\resizebox{0.9\linewidth}{!}{ 
\begin{tabular}{lrlrr}
\toprule
\textbf{Method} & \textbf{BLEU} & \textbf{COMET} & \textbf{CHRF} & \textbf{BERT} \\
\midrule
\textbf{TreePrompt-324+AFSP}   & 0.0293 & \textbf{-0.1475} & 28.33 & 0.9264 \\
\textbf{AFSP}            &\textbf{ 0.0353} & -0.1581 & \textbf{28.52} & 0.9265 \\
TreePrompt-324+KNN+Rerank & 0.0339 & -0.1758 & 28.15 & 0.9263 \\
KNN+Rerank      & 0.0329 & \-0.1760 & 28.56 & 0.9261 \\
\textbf{Random}          & 0.0308 & -0.1811 & 27.94 & \textbf{0.9266} \\
TreePrompt+Random+Rerank & 0.0350 & -0.1817 & 28.07 & 0.9262 \\
TreePrompt-144+AFSP   & 0.0282 & -0.1839 & 27.91 & 0.9261 \\
Random+Rerank      & 0.0263 & -0.1856 & 27.85 & 0.9259 \\
KNN             & 0.0326 & -0.1859 & 28.28 & 0.9261 \\
TreePrompt-144+KNN        & 0.0329 & -0.1860 & 28.04 & 0.9262 \\
TreePrompt-144+Random   & 0.0324 & -0.1963 & 27.88 & 0.9260 \\
TreePrompt-324+Random    & 0.0276 & -0.1980 & 28.34 & 0.9262 \\
Zero-shot       & 0.0251 & -0.2334 & 27.30 & 0.9259 \\
\bottomrule
\end{tabular}
}
\end{subfigure}

\vspace{0.5cm}

\begin{subfigure}[b]{\linewidth}
\centering
\caption{GPT-3.5 Turbo}
\resizebox{0.9\linewidth}{!}{
\begin{tabular}{lrlrr}
\toprule
\textbf{Method} & \textbf{BLEU} & \textbf{COMET} & \textbf{CHRF} & \textbf{BERT} \\
\midrule
\textbf{Random+Rerank}         & 0.0216 & \textbf{-0.3279} & 26.04 & 0.9239 \\
TreePrompt-596+KNN+Rerank    & 0.0204 & -0.3342 & 26.32 & 0.9237 \\
AFSP               & 0.0210 & -0.3361 & 25.98 & 0.9240 \\
\textbf{TreePrompt-596+AFSP}          & \textbf{0.0224} & -0.3437 & \textbf{26.46} & \textbf{0.9246} \\
TreePrompt-596+KNN           & 0.0189 & -0.3455 & 26.22 & 0.9235 \\
TreePrompt-596+Random+Rerank    & 0.0182 & -0.3459 & 26.04 & 0.9240 \\
TreePrompt-596+Random        & 0.0217 & -0.3522 & 26.16 & 0.9241 \\
Random             & 0.0124 & -0.3600 & 23.53 & 0.9232 \\
KNN+Rerank         & 0.0196 & -0.3636 & 26.34 & 0.9237 \\
KNN                & 0.0190 & -0.3653 & 26.30 & 0.9236 \\
Zero-shot          & 0.0193 & -0.3879 & 25.98 & 0.9243 \\
\bottomrule
\end{tabular}
}
\end{subfigure}

\vspace{0.5cm}

\begin{subfigure}[b]{\linewidth}
\centering
\caption{DeepSeek}
\resizebox{0.9\linewidth}{!}{
\begin{tabular}{lrlrr}
\toprule
\textbf{Method} & \textbf{BLEU} & \textbf{COMET} & \textbf{CHRF} & \textbf{BERT} \\
\midrule
\textbf{TreePrompt-653+Random+Rerank}  & 0.0358 & \textbf{-0.1424} & 28.57 & 0.9265 \\
TreePrompt-653+Random      & 0.0305 & -0.1492 & 28.65 & 0.9264 \\
\textbf{AFSP}             & \textbf{0.0402} & -0.1512 & \textbf{28.76} & 0.9264 \\
TreePrompt-653+KNN         & 0.0288 & -0.1579 & 28.30 & 0.9266 \\
TreePrompt-653+KNN+Rerank  & 0.0310 & -0.1617 & 28.32 & 0.9260 \\
Random           & 0.0325 & -0.1674 & 28.34 & 0.9264 \\
TreePrompt-653+AFSP        & 0.0354 & -0.1701 & 28.31 & 0.9263 \\
KNN              & 0.0291 & -0.1943 & 28.29 & 0.9256 \\
\textbf{Zero-shot}        & 0.0314 & -0.2037 & 28.71 & \textbf{0.9267} \\
\bottomrule
\end{tabular}
}
\end{subfigure}

\caption{Evaluation results for GPT-4o, GPT-3.5 Turbo, and DeepSeek across different prompting methods in the MIZAN dataset, English-Persian sorted by COMET score (primary metric) The highest scores and methods are bolded.}
\label{fig:llm-comet-compare}
\end{figure}

\subsection{Analysis of TreePrompt Method}

TreePrompt is computationally intensive because it involves a K-NN expansion algorithm followed by prompt labeling by the LLM. Table~\ref{tab:model_nneighboor_knn} summarizes the hyperparameters used in TreePrompt. The term Random refers to the number of randomly selected examples in the initial stage of the algorithm. N-neighbor denotes the K-NN parameter defining the number of nearest neighbors considered at each step, while K-NN indicates the total number of K-NN iterations applied to progressively select high-quality examples.

\begin{table}[ht]
\small
\setlength{\tabcolsep}{2pt}
\renewcommand{\arraystretch}{1.2}
\centering
\begin{tabular}{|l|r|r|r|}
\hline
\textbf{Model} & \textbf{Random Examples} & \textbf{n-neighboor} & \textbf{ K-NN} \\
\hline
GPT-4o        & 200 & 220 & 10 \\
\hline
GPT3.5 Turbo  & 631 & 600 & 2 \\
\hline
DeepSeek& 631 & 600 & 4 \\
\hline
DeepSeek      & 50  & 50 & 16 \\
\hline
\end{tabular}
\caption{TreePrompt Hyperparameters and Number of K-NN Executions for Each LLM}
\label{tab:model_nneighboor_knn}
\end{table}
We can obtain the complete prompt sent to the LLM using the following formula:
\[
P_{\mathrm{TreePrompt}} = n_{\mathrm{nbr}} \times k_{\mathrm{NN}} + R
\]
\begin{align*}
n_{\mathrm{nbr}} & : \text{Number of neighbors in K-NN} \\
k_{\mathrm{NN}} & : \text{Number of K-NN executions} \\
R & : \text{Number of random examples}
\end{align*}

Based on this formula, we calculate the number of prompts generated for each LLM, as detailed in Table 3.

As shown in Table~\ref{tab:model_random_examples}, the performance of the TreePrompt method is highly sensitive to its parameter settings. Proper tuning of these parameters can significantly reduce computational costs. Specifically, decreasing the number of neighbors leads to more frequent K-NN executions but results in fewer total prompts. In contrast, increasing the number of neighbors and random examples reduces the frequency of K-NN runs while substantially increasing the number of prompts generated. Therefore, a balance must be struck between these parameters to optimize both performance and computational efficiency.

\begin{table}[t]
\centering
\begin{minipage}[t]{0.5\textwidth} 
\raggedleft
\small
\setlength{\tabcolsep}{3pt}
\renewcommand{\arraystretch}{1.2}
\begin{tabular}{lcccc}
\toprule
\textbf{} & \textbf{COMET} & \textbf{BLEU} & \textbf{CHRF} & \textbf{BERT} \\
\midrule
\textbf{K-NN} & \textbf{0.9004} & 0.4230 & 71.64 & 0.9444 \\
TreePrompt-554+Random+Rerank & 0.9003 & 0.4122 & 71.23 & 0.9428 \\
TreePrompt-554+Random & 0.8997 & 0.4171 & 71.41 & 0.9431 \\
\textbf{Zero-shot} & 0.8997 & \textbf{0.4419} & \textbf{73.31} & \textbf{0.9458} \\
Random & 0.8995 & 0.4145 & 71.35 & 0.9432 \\
TreePrompt-554+KNN & 0.8994 & 0.4142 & 71.33 & 0.9436 \\
AFSP & 0.8993 & 0.4143 & 71.40 & 0.9437 \\
TreePrompt-554+AFSP & 0.8987 & 0.4144 & 71.25 & 0.9423 \\
\bottomrule
\end{tabular}
\caption{The results of English-German in WMT 19 datasets on DeepSeek. The highest scores and methods are bolded.}
\label{tab:English-German results}
\end{minipage}
\end{table}

\begin{table}[ht]
\small
\setlength{\tabcolsep}{2pt}
\renewcommand{\arraystretch}{1.2}
\centering
\begin{tabular}{|l|r|}
\hline
\textbf{Model} & \textbf{number of Prompts} \\
\hline
GPT-4o        & 2,400 \\
\hline
GPT3.5 Turbo  & 1,831 \\
\hline
DeepSeek& 3,031 \\
\hline
DeepSeek      & 850  \\
\hline
\end{tabular}
\caption{The number of prompts in TreePrompt for each LLM}
\label{tab:model_random_examples}
\end{table}

In summary, TreePrompt incurs higher computational costs but achieves more efficient translation by using fewer, higher-quality examples. It demonstrates that incorporating LLM preferences is crucial for selecting effective translation prompts. By integrating a tree structure with LLM feedback, TreePrompt offers valuable insights into identifying the most promising examples for few-shot machine translation.

\section{ Conclusion}
In this work, we introduced TreePrompt, a hierarchical and adaptive few-shot example selection method for machine translation with large language models (LLMs). TreePrompt leverages LLM-based quality scoring combined with semantic expansion through KNN-guided retrieval over RoBERTa embeddings to construct a structured and contextually relevant prompt set. We evaluated TreePrompt against several prompting strategies on three LLMs (GPT-4o, GPT-3.5-Turbo, and DeepSeek) using the MIZAN English–Persian parallel corpus and the WMT19 English–German dataset with one LLM.

Our English–Persian translation experiments demonstrate that combining TreePrompt with random selection and AFSP yields robust performance. For English–German translation, TreePrompt achieves competitive results compared to K-NN. Notably, our findings reveal that comparable translation quality can be attained using fewer, higher-quality examples rather than a large number of prompts. TreePrompt effectively identifies high-quality examples that strongly align with LLM preferences.

While TreePrompt is computationally intensive, appropriate parameter tuning allows it to efficiently discover good examples early in the selection process.

\section*{Limitations}
This study provides an initial exploration of the TreePrompt example selection method using the MIZAN and WMT19 datasets. While the results are promising across different prompting strategies and LLMs, the relatively small test set limits the generalizability of our conclusions.

Negative COMET scores were observed in some cases—particularly for English–Persian—which is expected for low-resource language pairs. Because COMET is primarily trained on high-resource languages, it may undervalue semantically correct translations that diverge from limited or literal references. Thus, while COMET offers useful insights, it may not fully capture translation quality in low-resource settings.

Future work should test TreePrompt on larger and more diverse corpora, spanning additional language pairs. Incorporating human evaluation will also be essential to better assess adequacy, fluency, and contextual relevance—dimensions that automated metrics alone cannot fully measure.

\section*{Acknowledgments}

We used generative AI tools (ChatGPT) solely for language polishing and improving the clarity of the manuscript. No new content, ideas, or analyses were generated by the AI. All research contributions and the final responsibility for the work rest with the authors.
\bibliography{anthology,custom}
\bibliographystyle{acl_natbib}

\appendix
\section{Appendix A}
\label{sec:appendix}
\subsection{ Example Translations from English to German (WMT’19 Dataset)}

This appendix showcases selected English sentences from the WMT 2019 dataset along with their German translations, serving as examples to illustrate and compare the quality of outputs produced by different methods examined in this work: Random Few-Shot prompting, K-Nearest Neighbors (KNN), Adaptive Few-Shot Prompting (AFSP), and TreePrompt. For each example, we present the source English sentence, the human reference translation, and the corresponding outputs from each method. These qualitative examples complement the quantitative results discussed in the main text.
\subsection{ source and reference}
\begin{itemize}
 \item \textbf{source} Only the dwarf tower that is still present today survived as witness of the church's 750-year-old history.
    \item \textbf{refrence} Nur der heute noch vorhandene Turmzwerg blieb als Zeuge der 750 Jahre alten Kirchengeschichte erhalten.
    \item \textbf{source} Up until 1869, the Hilarius cemetery in Tumlingen served as the last place of rest for Hörschweiler's dead.
    \item \textbf{refrence}Bis 1869 diente der Tumlinger Hilarius-Friedhof auch als letzte Ruhestätte für die Toten aus Hörschweiler.
    \item \textbf{source}During the 30 Years War (1616 to 1648) the window panes were smashed in and the bell ropes stolen.
    \item \textbf{refrence}Im 30-jährigen Krieg (1616 bis 1648) wurden die Fensterscheiben eingeschlagen und die Glockenseile gestohlen.
    \item \textbf{source}As is apparent from the upper administrative office for Freudenstadt's description in 1858, the \"mountain\" was, alongside the lower village and the upper village that can just about be made out still in the appearance of Tumlingen today, the village's third, grand centre of settlement.",
   \item \textbf{refrence}Wie aus der Beschreibung des Oberamtes Freudenstadt von 1858 hervorgeht, war neben dem Unterdorf und dem Oberdorf, die heute im Ortsbild von Tumlingen gerade noch zu erkennen sind, der \"Berg\" der dritte, der herrschaftliche Siedlungskern des Dorfes.
   \item \textbf{source}Due to its idyllic setting and beautiful view over the village, the historical site of St. Hilarius is a beloved place to tarry a while.
    \item \textbf{refrence}Ob seiner idyllischen Lage und schöner Aussicht auf die Ortschaft ist die historische St. Hilarius-Anlage ein beliebter Ort zum Verweilen.
\end{itemize}
   
\subsection{Zero-shot Method Translation}
\begin{itemize}
    \item Nur der Zwergturm, der noch heute vorhanden ist, überlebte als Zeuge der 750-jährigen Geschichte der Kirche.
    \item Bis 1869 diente der Hilarius-Friedhof in Tumlingen als letzte Ruhestätte für die Toten von Hörschweiler.
    \item Während des Dreißigjährigen Krieges (1616 bis 1648) wurden die Fensterscheiben eingeschlagen und die Glockenseile gestohlen..
    \item Wie aus der Beschreibung des oberen Verwaltungsamtes für Freudenstadt von 1858 hervorgeht, war der \"Berg\" neben dem unteren Dorf und dem oberen Dorf, das in der heutigen Erscheinung von Tumlingen noch erkennbar ist, das dritte, große Siedlungszentrum des Dorfes.
    \item Aufgrund seiner idyllischen Lage und des schönen Ausblicks über das Dorf ist die historische Stätte von St. Hilarius ein beliebter Ort, um eine Weile zu verweilen.
\end{itemize}

\subsection{Random Method Translation}
\begin{itemize}
    \item Nur der heute noch vorhandene Zwergturm überdauerte als Zeuge der 750-jährigen Geschichte der Kirche.
    \item Bis 1869 diente der Hilarius-Friedhof in Tumlingen als letzte Ruhestätte für die Toten aus Hörschweiler.
    \item Im Dreißigjährigen Krieg (1616 bis 1648) wurden die Fensterscheiben eingeschlagen und die Glockenseile gestohlen.
    \item Wie aus der Beschreibung des Oberamts Freudenstadt von 1858 hervorgeht, war der \"Berg\" neben dem unteren Dorf und dem oberen Dorf, das im heutigen Erscheinungsbild Tumlingens noch gerade zu erkennen ist, das dritte große Siedlungszentrum des Dorfes.
    \item Aufgrund seiner idyllischen Lage und der schönen Aussicht über das Dorf ist der historische Ort St. Hilarius ein beliebter Platz, um eine Weile zu verweilen.
\end{itemize}

\subsection{K-NN Method Translation}
\begin{itemize}
    \item Nur der heute noch vorhandene Zwergturm überdauerte als Zeuge der 750-jährigen Geschichte der Kirche.
    \item Bis zum Jahr 1869 diente der Hilarius-Friedhof in Tumlingen als letzte Ruhestätte für die Toten aus Hörschweiler.
    \item Während des Dreißigjährigen Krieges (1616 bis 1648) wurden die Fensterscheiben eingeschlagen und die Glockenseile gestohlen.
    \item Wie aus der Beschreibung des Oberamts Freudenstadt von 1858 hervorgeht, war der „Berg“ neben dem unteren Dorf und dem oberen Dorf, das im heutigen Erscheinungsbild Tumlingens noch gerade eben auszumachen ist, das dritte, große Siedlungszentrum des Ortes.
    \item Aufgrund seiner idyllischen Lage und der schönen Aussicht über das Dorf ist die historische Stätte St. Hilarius ein beliebter Ort, um eine Weile zu verweilen.
    
\end{itemize}

\subsection{AFSP Method Translation}
\begin{itemize}
    \item Nur der heute noch vorhandene Zwergturm überdauerte als Zeuge der 750-jährigen Geschichte der Kirche.
    \item Bis 1869 diente der Hilarius-Friedhof in Tumlingen als letzte Ruhestätte für die Toten von Hörschweiler.
    \item Im Dreißigjährigen Krieg (1616 bis 1648) wurden die Fensterscheiben eingeschlagen und die Glockenseile gestohlen.
    \item Wie aus der Beschreibung des Oberamts Freudenstadt von 1858 hervorgeht, war der \"Berg\" neben dem unteren Dorf und dem oberen Dorf, das im heutigen Erscheinungsbild Tumlingens noch gerade erkennbar ist, das dritte große Siedlungszentrum des Dorfes.
    \item Aufgrund seiner idyllischen Lage und der schönen Aussicht über das Dorf ist der historische Ort St. Hilarius ein beliebter Platz, um eine Weile zu verweilen."
\end{itemize}

\subsection{TreePrompt-554+Random Method Translation}
\begin{itemize}
    \item Nur der heute noch vorhandene Zwergturm überdauerte als Zeuge der 750-jährigen Geschichte der Kirche.
    \item Bis 1869 diente der Hilarius-Friedhof in Tumlingen als letzte Ruhestätte für die Toten aus Hörschweiler.
    \item Während des Dreißigjährigen Krieges (1616 bis 1648) wurden die Fensterscheiben eingeschlagen und die Glockenseile gestohlen.
    \item Wie aus der Beschreibung des Oberamts Freudenstadt von 1858 hervorgeht, war der \"Berg\" neben dem unteren Dorf und dem oberen Dorf, das im heutigen Erscheinungsbild Tumlingens noch gerade eben auszumachen ist, das dritte große Siedlungszentrum des Ortes.
    \item Aufgrund seiner idyllischen Lage und der schönen Aussicht über das Dorf ist der historische Ort St. Hilarius ein beliebter Platz, um eine Weile zu verweilen.
\end{itemize}

\subsection{TreePrompt-554+Random+Rerank Method Translation}
\begin{itemize}
    \item Nur der heute noch vorhandene Zwergturm überdauerte als Zeuge der 750-jährigen Geschichte der Kirche.
    \item Bis zum Jahr 1869 diente der Hilarius-Friedhof in Tumlingen als letzte Ruhestätte für die Toten aus Hörschweiler.
    \item Während des Dreißigjährigen Krieges (1616 bis 1648) wurden die Fensterscheiben eingeschlagen und die Glockenseile gestohlen.
    \item Wie aus der Beschreibung des Oberamts Freudenstadt von 1858 hervorgeht, war der \"Berg\" neben dem unteren Dorf und dem oberen Dorf, das im heutigen Erscheinungsbild Tumlingens noch gerade erkennbar ist, das dritte große Siedlungszentrum des Dorfes.
    \item Aufgrund seiner idyllischen Lage und der schönen Aussicht über das Dorf ist der historische Ort St. Hilarius ein beliebter Platz, um eine Weile zu verweilen.
\end{itemize}

\subsection{TreePrompt-554+AFSP Method Translation}
\begin{itemize}
    \item Nur der heute noch vorhandene Zwergturm überdauerte als Zeuge der 750-jährigen Geschichte der Kirche.
    \item Bis zum Jahr 1869 diente der Hilarius-Friedhof in Tumlingen als letzte Ruhestätte für die Verstorbenen aus Hörschweiler.
    \item Während des Dreißigjährigen Krieges (1616 bis 1648) wurden die Fensterscheiben eingeschlagen und die Glockenseile gestohlen.
    \item Wie aus der Beschreibung des Oberamts Freudenstadt von 1858 hervorgeht, war der \"Berg\" neben dem unteren Dorf und dem oberen Dorf, das im heutigen Erscheinungsbild Tumlingens noch gerade erkennbar ist, das dritte große Siedlungszentrum des Ortes.
    \item Aufgrund seiner idyllischen Lage und der schönen Aussicht über das Dorf ist die historische Stätte St. Hilarius ein beliebter Ort, um eine Weile zu verweilen.
\end{itemize}

\subsection{TreePrompt-554+K-NN Method Translation}
\begin{itemize}
    \item Nur der heute noch vorhandene Zwergturm überdauerte als Zeuge der 750-jährigen Geschichte der Kirche.
    \item Bis 1869 diente der Hilarius-Friedhof in Tumlingen als letzte Ruhestätte für die Toten aus Hörschweiler.
    \item Während des Dreißigjährigen Krieges (1616 bis 1648) wurden die Fensterscheiben eingeschlagen und die Glockenseile gestohlen.
    \item Wie aus der Beschreibung des Oberamts Freudenstadt von 1858 hervorgeht, war der \"Berg\" neben dem unteren Dorf und dem oberen Dorf, das im heutigen Erscheinungsbild Tumlingens noch gerade erkennbar ist, das dritte große Siedlungszentrum des Dorfes.
    \item Aufgrund seiner idyllischen Lage und der schönen Aussicht über das Dorf ist der historische Ort St. Hilarius ein beliebter Platz, um eine Weile zu verweilen.
\end{itemize}

\end{document}